%% file: main.tex
\documentclass[10pt,twocolumn,letterpaper]{article}

\usepackage[pagenumbers]{cvpr} 
\usepackage{graphicx}   
\usepackage{booktabs}   
\usepackage{stfloats}
\usepackage{multirow}

\input{preamble}
\definecolor{cvprblue}{rgb}{0.21,0.49,0.74}
\usepackage[pagebackref,breaklinks,colorlinks,allcolors=cvprblue]{hyperref}

\def\paperID{9149} 
\def\confName{CVPR}
\def\confYear{2026}

\title{SciTextures: Collecting and Connecting Visual Patterns, Models, and Code Across Science and Art}

\author{Sagi Eppel\textsuperscript{1}, Alona Strugatski\textsuperscript{1}
}

\begin{document}
\maketitle

\input{sec/0_abstract}

\input{sec/1_intro}
\input{sec/related_work}

\input{sec/Dataset}

\input{sec/table}
\input{sec/evaluation_tasks}
\input{sec/result_and_conclusion}
{
    \small
    \bibliographystyle{ieeenat_fullname}
    \bibliography{main}
}


\end{document}

%% file: sec/0_abstract.tex
\begin{figure*}[b]
  \centering
  \vspace{-10pt}
   \includegraphics[width=1\textwidth]{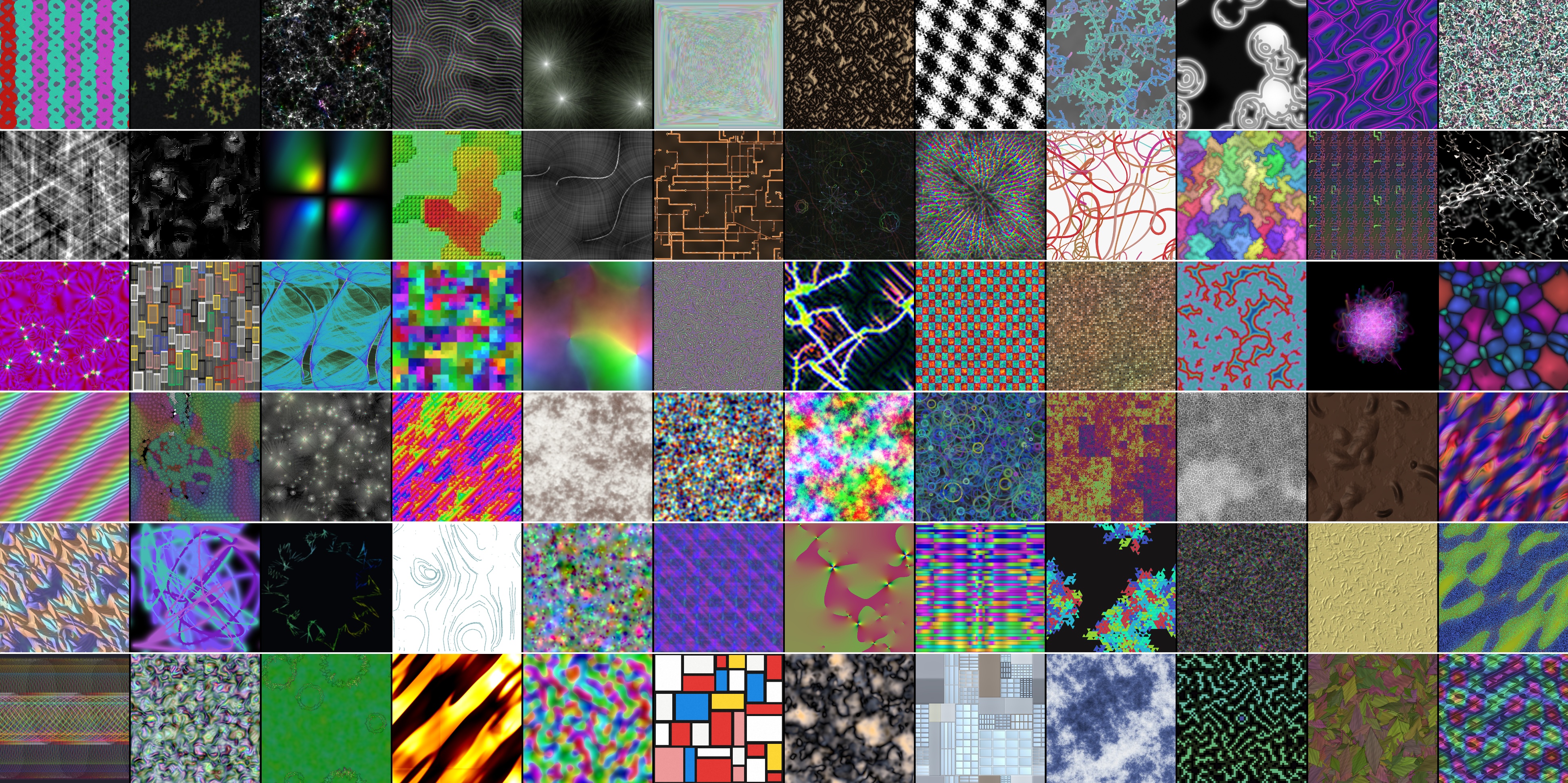}
  
  \caption{The SciTextures is a large-scale collection of textures and visual patterns, combined with the code (model, simulation, or method) that produced it. The dataset contains over 100,000 images from 1,270 systems, spanning diverse domains of science and art.
}
  \label{fig:fig1}
\end{figure*}
\begin{abstract}
The ability to connect visual patterns with the processes that form them represents one of the deepest forms of visual understanding. Textures of clouds and waves, the growth of cities and forests, or the formation of materials and landscapes are all examples of patterns emerging from underlying mechanisms. We present the SciTextures dataset, a large-scale collection of textures and visual patterns from all domains of science, tech, and art, along with the models and code that generate these images. Covering over 1,270 different models and 100,000 images of patterns and textures from physics, chemistry, biology, sociology, technology, mathematics, and art, this dataset offers a way to explore the deep connection between the visual patterns that shape our world and the mechanisms that produce them. 
Built through an agentic AI pipeline that autonomously collects, implements, and standardizes scientific and generative models. This AI pipeline is also used to autonomously invent and implement novel methods for generating visual patterns and textures. SciTextures enables systematic evaluation of AI’s ability to link visual patterns to the models and code that generate them, and to identify different patterns that emerge from the same underlying process. We also test vision-language-models (VLM's) ability to infer and recreate the mechanisms behind visual patterns by providing a natural image of a real-world phenomenon and asking the AI to identify and code a model of the process that formed it, then run this code to generate a simulated image that is compared to the reference image. These benchmarks reveal that VLM's can understand and simulate physical systems beyond visual patterns at multiple levels of abstraction. The dataset and code are available at \href{https://sites.google.com/view/scitextures/home}{1},\href{https://zenodo.org/records/17485502}{2}
\end{abstract}
\footnotetext{\textsuperscript{1}Weizmann Institute of Science, AI Hub}

%% file: sec/1_intro.tex
\begin{figure*}[b!]
  \centering
  \vspace{-10pt}
   \includegraphics[width=1\textwidth]{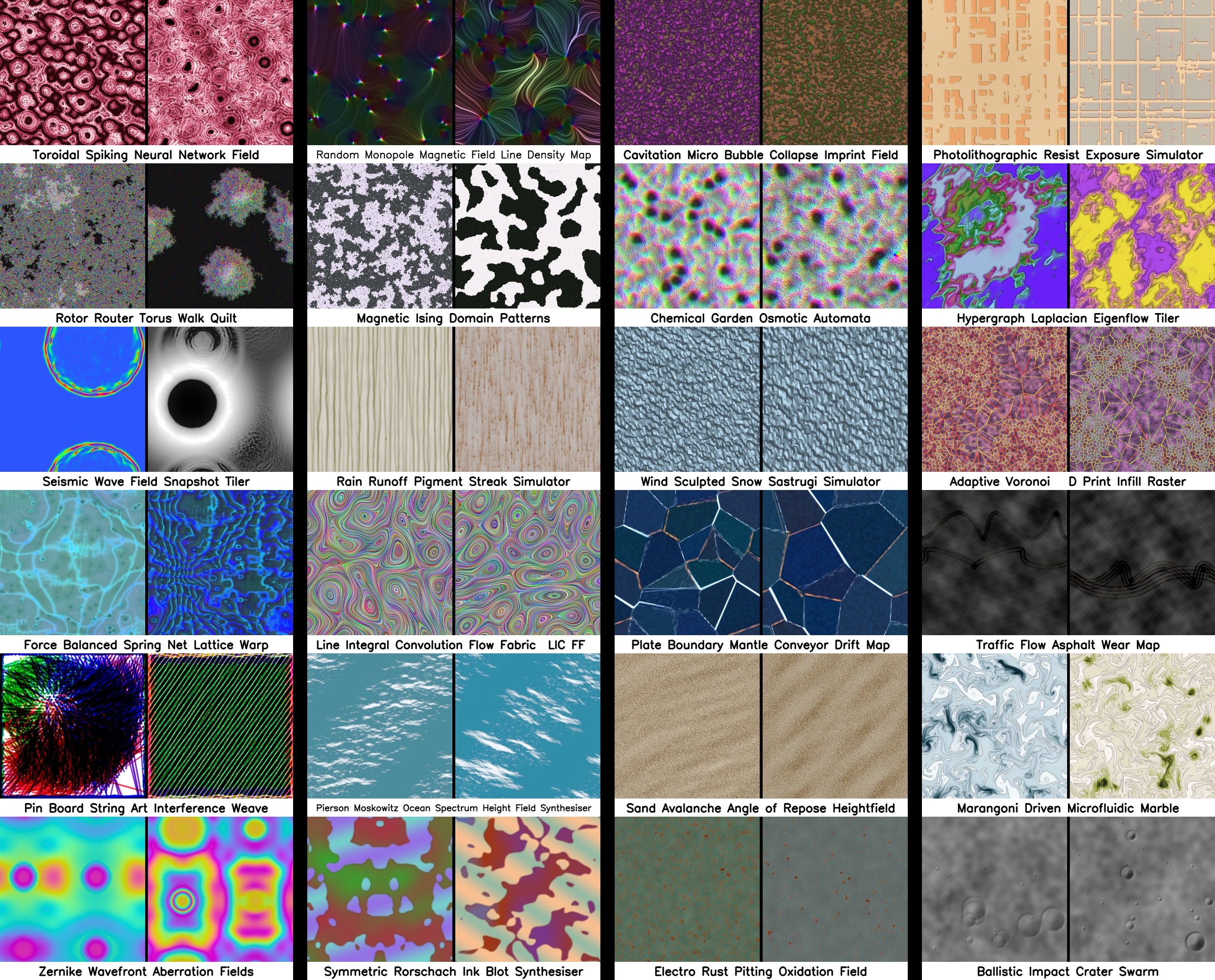}
  
  \caption{Sample image pairs from the SciTextures dataset. Each horizontal pair was generated by the same model, with the model’s name shown below. The dataset includes 80 images per model, covering 1,270 models and their corresponding code implementations.
}
  \label{fig:fig2}
\end{figure*}
\section{Introduction}
\label{sec:intro}
\begin{figure*} 
  \centering
   \includegraphics[width=1\textwidth]{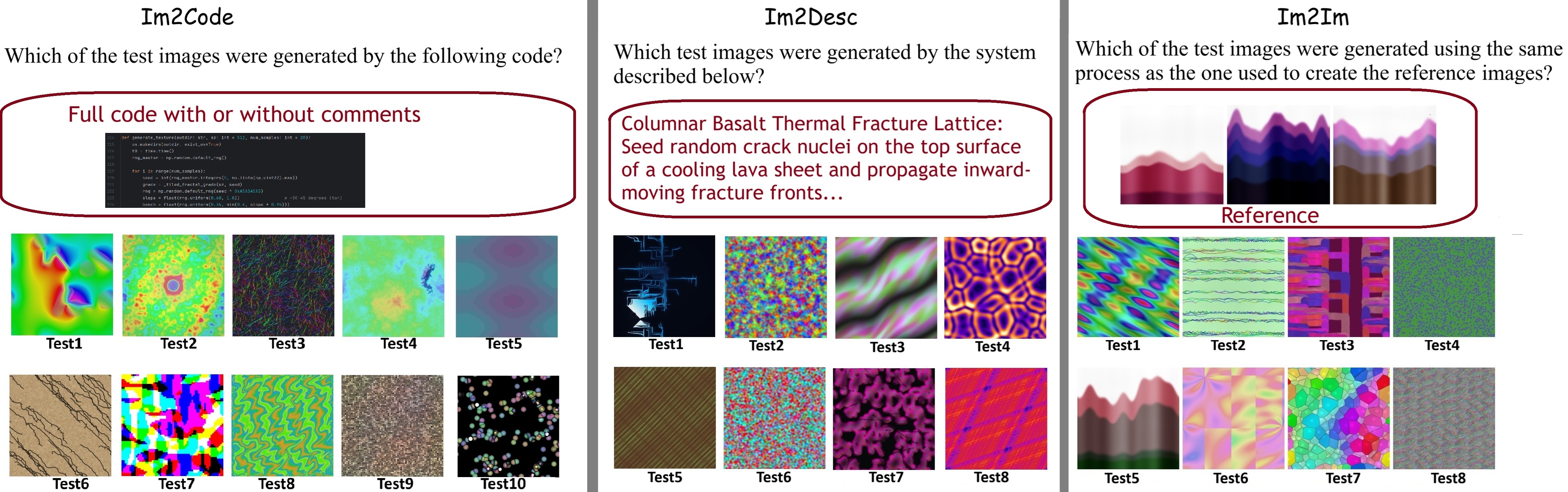}
  
  \caption{Different approaches for evaluating AI’s ability to connect visual patterns with the underlying process that generated them. \textbf{Im2Code:} The AI is given a piece of code (with or without comments) and several images, and must identify which image(s) were generated by that code. \textbf{Im2Desc:} The AI receives a textual description of a system along with several images produced by different systems, and must determine which image(s) were generated by the system described. \textbf{Im2Im:} The AI is given a few reference images generated by the same model, along with several test images, and must identify which test image(s) were created by the same process as the reference images.
}
  \label{fig:fig3}
\end{figure*}

\begin{figure*}[b!]
  \centering
   \includegraphics[width=1\textwidth]{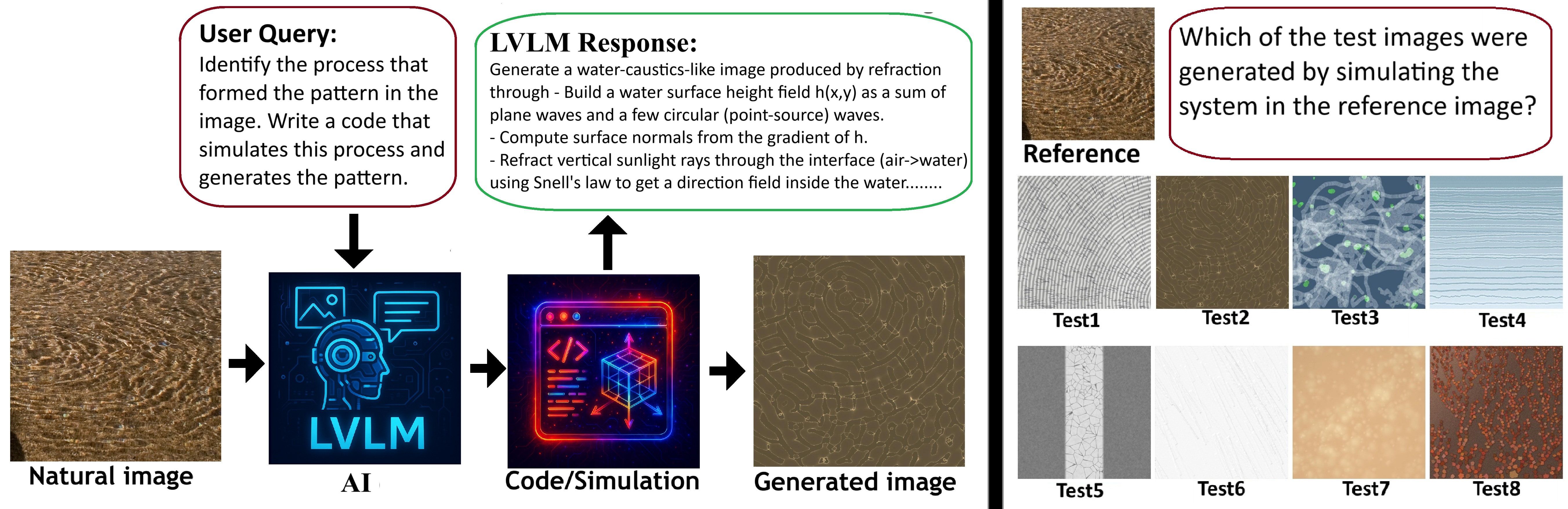}
  
  \caption{The \textbf{Im2Sim2Im} approach for testing an AI’s ability to identify and model the underlying process behind a real-world visual pattern. \textbf{Left:} The AI receives an image of a real-world pattern, infers the physical process that formed this pattern, implements it as code, and runs the code to generate a simulated image. \textbf{Right:} A matcher evaluates and ranks the simulated image by identifying which test image best matches the reference image containing the real-world pattern. The hypothesis is that if the matching process (right) is accurate, it can be indirectly used to evaluate the quality of the generated simulation (left).
}
  \label{fig:fig4}
\end{figure*}

The visual world is a tapestry of textures and visual patterns emerging from the underlying physical reality\cite{ball1998self,turing1990chemical,brent1978prigogine,thomson1917growth,cross1993pattern,wolfram1997new,hofstadter1999godel,mandelbrot1983fractal,flake2000computational,lague_youtube,emergentgarden_youtube,sims1991artificial,sims1991artificial}. The ability to look at a visual pattern and infer the process that formed it represents one of the deepest forms of visual understanding\cite{gregory2015eye,rock1983logic}.  Every visual pattern arises from an underlying mechanism, from the self-assembly of materials to the growth of crystals, colonies, and ecosystems, and even the formation of landscapes, plants, and galaxies. Each of these spatial patterns is a visible trace of the hidden dynamics that generate them. Understanding and modeling how such patterns arise and how they connect to the mechanisms that produce them is a central pursuit across science, mathematics, and visual art\cite{ball1998self,turing1990chemical,brent1978prigogine,thomson1917growth,cross1993pattern,wolfram1997new,hofstadter1999godel,mandelbrot1983fractal,flake2000computational,lague_youtube,emergentgarden_youtube,sims1991artificial,sims1991artificial}. Numerous works on this problem are scattered across a vast range of domains (\cref{fig:fig1,fig:fig2}). While some works have sought to treat visual patterns emergence more generally\cite{ball1998self,turing1990chemical,brent1978prigogine,thomson1917growth,cross1993pattern,wolfram1997new,hofstadter1999godel,mandelbrot1983fractal,flake2000computational,raistrick2023infinite,sims1991artificial}, to the best of our knowledge, no comprehensive dataset exists that collects these patterns and their corresponding models across a wide range of scientific domains.
In contrast, computer vision and AI researchers have developed various methods and datasets that explore texture and pattern perception without being limited to specific domains\cite{oquab2023dinov2,chen2020simple,drehwald2023one,cimpoi2014describing,sharma2023materialistic,raistrick2023infinite}. However, these approaches primarily focus on classifying or matching images or textures of the same type, while ignoring the underlying systems that generate them. As a result, they do not evaluate whether AI systems truly understand the processes behind an image. This work aims to address these limitations by introducing a large-scale, domain-general dataset that brings together visual patterns and textures from across science, technology, and art, alongside the models and code that simulate and generate them (\cref{fig:fig1,fig:fig2}). This was achieved through an automated agentic AI pipeline that collects, adapts, and implements models from a wide range of fields in a unified format. In addition, we apply this autonomous pipeline to invent and implement novel generative methods and simulations for textures and visual patterns generation.\newline
A second goal of this work is to probe the depth of Vision language models (VLM's) visual understanding by assessing their ability to identify and reconstruct the underlying generative processes behind visual patterns. To this end, we introduce three novel benchmarking tasks:  \newline
\textbf{a) Im2Code:}  Testing the VLM's ability to match images  of visual patterns to the code that generated them (\cref{fig:fig3}).\newline
\textbf{b) Im2Im:}  Assessing the VLM's ability to identify different images that originate from the same underlying process (\cref{fig:fig3}).\newline
\textbf{c) Im2Sim2Im:} Presenting an AI with a real-world image of a pattern, asking it to identify the process that produced it, write code to simulate that process, and run it to generate a synthetic image (\cref{fig:fig4}.left). The generated  image is then compared to the real-world input image to assess the model’s success (\cref{fig:fig4}.right).\newline   
Together, these tasks provide a general framework for evaluating whether AI models possess a deeper, mechanistic understanding of visual patterns.  Note that AI in this work refers to Vision Language Models (VLMs).

\subsection{Main Contributions}
a) This work introduces the first large-scale, general dataset that collects textures and visual patterns spanning multiple scientific domains, together with the models and code used to generate them. It enables data-driven exploration of visual patterns that transcend specific domains, linking visual form to underlying mechanisms and offering a new lens on one of science’s most universal phenomena.\newline
b) In addition, this work introduces the first autonomous pipeline that allows VLM's and LLM's to invent novel generative methods and simulations for textures and visual patterns. Offering an almost unlimited variety of code-based visual patterns.
c) Understanding a system is inherently tied to the ability to model it; likewise, identifying and modeling the mechanism behind an image reflects a deep form of visual understanding, one largely unexplored in computer vision. Along with the dataset, the evaluation methods presented here enable both quantitative and qualitative evaluation of this fundamental capability, and show that leading AIs can understand and model the processes beyond visual patterns across multiple levels of abstraction.

\begin{figure*} 
  \centering
   \includegraphics[width=1\textwidth]{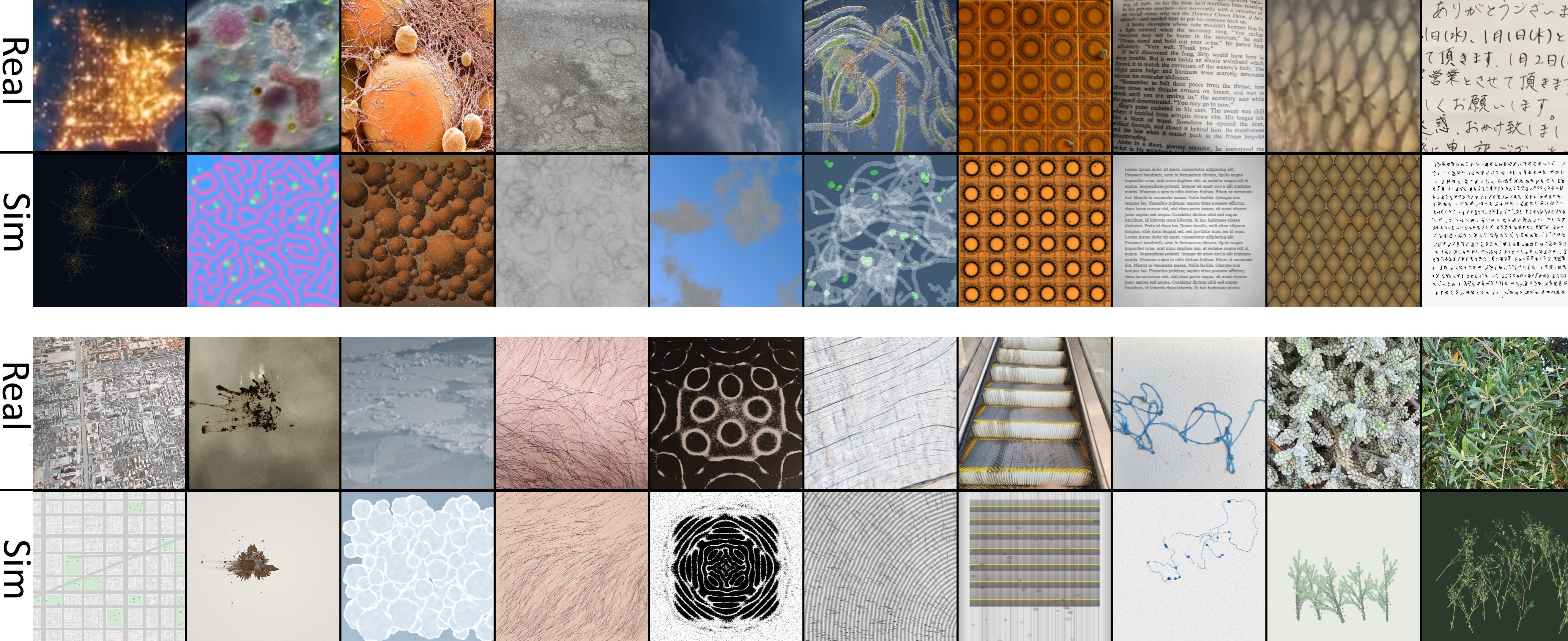}
  
  \caption{Result from the Im2Sim2Im (\cref{fig:fig4}, left). The “Real” image is a picture of a real-world pattern. The “Sim” image (below the real image) is the image generated by the code/simulation created by the AI in an attempt to model the system in the real image.
}
  \label{fig:fig5}
\end{figure*}

%% file: sec/related_work.tex
\section{Related Works}
\label{sec:related_work}

\textbf{Identifying visual patterns and the processes} that generate them is central to both science and visual art\cite{ball1998self,turing1990chemical,brent1978prigogine,thomson1917growth,cross1993pattern,wolfram1997new,hofstadter1999godel,mandelbrot1983fractal,flake2000computational,lague_youtube,emergentgarden_youtube,raistrick2023infinite,sims1991artificial}. While numerous models, simulations, and graphics techniques exist for these tasks, they remain scattered across disparate fields. Few works have explored this territory in a more unified way\cite{thomson1917growth,cross1993pattern,wolfram1997new,hofstadter1999godel,mandelbrot1983fractal,flake2000computational}, often focusing on common mathematical principles like fractals\cite{mandelbrot1983fractal,flake2000computational}, cellular automata\cite{wolfram1997new}, self-organization and emergence\cite{turing1990chemical,cross1993pattern,thomson1917growth} that universally recur across different systems.\newline
\textbf{Datasets of visual patterns and textures} typically focus on specific phenomena and rarely include generation models. The largest source of linked visual patterns and executable scripts comes from OpenGL/WebGL shaders: short web scripts used to generate real-time graphics and simple videos in websites. The most extensive collection of such scripts is ShaderToy\cite{jeremias2014shadertoy,baradad2022procedural}, containing over 51,000 shaders that can each generate video output. However, these shaders are designed for CGI and artistic projects with real-time rendering capabilities (~979 frames per second\cite{baradad2022procedural}), not for scientific models and simulations, which typically cannot run in real time and prioritize physical accuracy over visual aesthetics. Another source of texture-program pairs comes from node-graph systems for PBR materials and shaders, though these remain art-focused and limited to surface textures\cite{blenderkit}. Scientific datasets that combine code and patterns are typically domain-specific, concentrating on areas like fractals\cite{hirokatsu2022pre}, metamaterials\cite{makatura2025metagen}, morphogenesis\cite{lomas2014cellular,lomas2016species}, or procedurally generated training data\cite{cobbe2020leveraging,bray2012workflow, raistrick2023infinite}.\newline
\textbf{This work also relates to visual program synthesis}, which involves inferring underlying programs from images. Works in this field focused on extracting simple code from  technical diagrams such as CAD, SVG, Webpages, or code diagrams\cite{pearl2025geocode,reddy2021im2vec,du2018inversecsg,wu2021deepcad,beltramelli2018pix2code,doris2025cad,ren2025simugen,li2025vlmaterial}. While early approaches used specialized neural networks\cite{carlier2020deepsvg,reddy2021im2vec,du2018inversecsg,wu2021deepcad}, recent work has begun exploring foundation vision-language models (VLMs)\cite{li2025vlmaterial}. One example task involves reconstructing the node-graph that generated a given texture image\cite{li2025vlmaterial,hu2023generating,guerrero2022matformer} (node-graphs being a form of visual scripting used to create textures). An older but related approach is exemplar-based texture synthesis, which can be viewed as simplified model inference from images\cite{raad2017survey,mumford2010pattern,liu2018perception}. This methods focuses on identifying the statistical building blocks of a pattern or finding system parameters that replicate it.

%% file: sec/Dataset.tex
\section{The SciTextures Dataset  }
\label{sec:dataset}

The SciTextures dataset is a large-scale collection of images featuring visual patterns and textures from a wide range of scientific and artistic domains, along with the corresponding models and generation code that produce them (\cref{fig:fig1,fig:fig2}). The dataset spans over 1,270 distinct generative models, ranging from simple systems and mathematical functions, such as the Ising model and the Game of Life, to simulations of cities, materials, chemical reactions, biological growth, and many others. Each model is accompanied by standardized, flexible code that generates a given number of images at arbitrary resolutions, as well as 80 images produced by the model. Most images are colored and seamlessly tileable, achieved through the use of periodic boundary conditions. In total, the dataset includes over 100,000 images, 1,270 models, and around 300,000 lines of code.
\subsection{Dataset Generation: Using Agentic AI to Suggest and Implement Models}
Identifying visual patterns and the mechanisms that form them is a central pursuit in both science and art. However, the vast number of methods developed for this purpose makes manual collection and implementation impractical. To overcome this challenge, we introduce an agentic AI pipeline that autonomously collects, implements, and validates these models, requiring human involvement only during the review stage.
The model gathering process begins with an AI agent proposing a set of models for visual pattern generation. This can be done either from existing scientific and artistic principles, using  novel ideas suggested by the LLM, or  by using VLM to try replicate the pattern in a natural image.  Each model is implemented as executable code, debugged, and reviewed by multiple AI systems to ensure conceptual soundness. The models are then run to generate simulated images, which are manually screened for quality and diversity. A manual audit of a few sampled scripts confirmed the reliability of the automated modeling pipeline. All generation scripts and prompts are supplied in the S.I.  More details for each step are presented below: \newline
\textbf{1) Model Suggestion:} The AI agent (GPT-5) is tasked with proposing a set of models capable of generating visual patterns (five ideas per round) based on either established methods, like published models and simulations from any domain of science, tech, and art,  or original concepts suggested by the LLM. Alternatively, A VLM is provided with a natural image of a real-world pattern (\cref{fig:fig4}) and asked to infer and model the process that produced it. The agent has access to all previously generated models in the dataset and is asked to avoid and filter repeating ideas, and suggest models from as many fields as possible.\newline
\textbf{2) Code Implementation:} The AI agent is then instructed to independently implement each proposed model as standardized code that generates any number of images at variable resolutions.\newline
\textbf{3) Inspection and Debugging:} The generated code is executed, and both the outputs and source code undergo multiple rounds of inspection and debugging. Even when the code runs successfully, additional reviews are performed by multiple AI agents (GPT-5, DeepSeek R1, and Claude Sonnet 4.5) to detect potential conceptual or logical errors.\newline
\textbf{4) Image Inspection:} Few images from each model are inspected both manually and automatically to identify problematic outputs, such as pure noise (manually), uniform images (automatic), or models that produce repetitive, overly similar results (model must be capable of generating an unlimited variety of distinct images).\newline
\textbf{5) Model Accuracy Ranking:} Each model is evaluated by additional AI agents (Claude Sonnet 4.5 and DeepSeek R1), ranking how accurately the model simulates the target system (See \cref{sec:model_fidelity} for details). \newline
\textbf{6) Manual Code Inspection:} Since manually inspecting 300,000 lines of code is not realistic, we have sampled 50 models with which we are familiar and manually inspected them, and could not find any significant errors.

\subsection{Model Fidelity and Granularity}
\label{sec:model_fidelity}
The degree to which a model faithfully represents the underlying mechanism of a system can differ widely. For example, ocean waves can be simulated using complex fluid and wind dynamics or as a simple sinusoidal function. Each model in the dataset was evaluated and ranked with five categories of fidelity by Claude 4.5 Sonnet and DeepSeek R1. The ranking categories are as follows: \textbf{1) Accurate (4\%):} Models that faithfully and quantitatively reproduce the system.  \textbf{2) Good Approximation (42\%):} Models that capture the system’s behavior with reasonable accuracy, though not exhaustively. \textbf{3) Toy Models (40\%):}  Simplified models that capture the core qualitative behavior but lack quantitative fidelity. \textbf{4) Weak Approximations (1\%):} Models that represent only minor aspects of the system, missing most critical dynamics. \textbf{5) Inspired By (13\%):}  Models, often artistic/graphics, that aim to replicate the system’s appearance rather than its underlying mechanisms.

\subsection{Prompts and Creativity}
We used multiple sets of prompts to guide the AI pipeline; some sets direct the AI toward suggesting established models, while others encourage it to generate novel ideas. The AI identified models across a wide range of disciplines, surpassing, by far, the knowledge of any single person. Even for canonical models, substantial variability exists in parameters, visualization choices,  and the projection of patterns into images, providing room for creativity in implementation. When prompted to prioritize creativity, the AI frequently leverages known functions and models, recombining them in novel configurations. In most cases, the models produced colorful images and seamless, tileable patterns (with periodic boundaries), making the results valuable as visual art assets. All prompts are supplied together with the generation code in the S.I.

%% file: sec/table.tex
\begin{table*}[t]
\centering

\caption{Results of the various tests (all values in \%). Each test consisted of 100 questions with 10 possible choices per question, yielding a random-guess baseline of 10\%. \textbf{Im2Desc}  (\cref{fig:fig3}): matching the model description (text) to a set of images (matching one text description to images from 10 different models, with 3 reference  images from each model). \textbf{Im2Code} (\cref{fig:fig3}): Identifying which image was generated by a given code, for code with or without \textbf{comments} (Same format as Im2Desc). \textbf{Im2Im} (\cref{fig:fig3}): Identifying which images were generated by the same model (With 3 reference images generated by one model and 10 test images each generated by a different model). \textbf{Im2Sim2Im} (\cref{fig:fig4}): The VLM is given a natural image and is asked to identify the process that generates the pattern in the image, model it, and generate a simulated image. The matching accuracy of the simulated image  to the input natural image is used to evaluate the model accuracy (\cref{fig:fig4}).  Matching was done with a single real image as a reference and 10 synthetic test images, each made by a different model. The image matching/ranking (\cref{fig:fig4}.left) of Im2Sim2Im was done using using GPT-5 with \textbf{color} or \textbf{gray}scale images, or by a \textbf{human} evaluator (with color images).}
\label{tab:results}
\small
\begin{tabular}{l@{\hspace{4pt}}|c@{\hspace{4pt}}c@{\hspace{4pt}}c@{\hspace{4pt}}c@{\hspace{4pt}}c@{\hspace{4pt}}c@{\hspace{4pt}}|c@{\hspace{4pt}}c@{\hspace{4pt}}c}
\toprule
& \multicolumn{6}{c}{Task Performance (\%)} & \multicolumn{3}{|c}{Im2Sim2Im} \\
\cmidrule(lr){2-7} \cmidrule(lr){8-10}
\scriptsize Model & \scriptsize Im2Desc & \scriptsize \shortstack{Im2Code\\Comments} & \scriptsize \shortstack{Im2Code\\Clean} & \scriptsize Im2Model2Code & \scriptsize Im2Model2Im & \scriptsize Im2Im & \scriptsize Color & \scriptsize Gray & \scriptsize Human \\
\midrule
GPT-5 & 44 & 58 & 50 & 38 & 48 & 95 & 74 & 62 & 82 \\
GPT-5-mini & 43 & 50 & 49 & 43 & 50 & 94 & 77 & 70 & 75 \\
Gemini-2.5-flash & 44 & 44 & 51 & 33 & 47 & 94 & 79 & 64 & 74 \\
Gemini-2.5-pro & 50 & 50 & 55 & 35 & 38 & 92 & 78 & 67 & 68 \\
Qwen2.5-VL-72B & 29 & 37 & 30 & 20 & 18 & 73 & 41 & 35 & 45 \\
Llama-4-Maverick-17B & 38 & 42 & 39 & 20 & 28 & 87 & 54 & 47 & 60 \\
Gemma-3n-E4B-it & 11 & 8 & 13 & 13 & 22 & 52 & -- & -- & -- \\
Llama-4-Scout-17B & 32 & 27 & 32 & 14 & 25 & 85 & 54 & 46 & 61 \\
Grok-4-fast-reasoning & 13 & 14 & 14 & 11 & 21 & 47 & 62 & 62 & 70 \\
Grok-4-fast-non-reasoning & 12 & 10 & 13 & 12 & 14 & 36 & 56 & 48 & 55 \\
Grok-4 & 13 & 10 & 14 & 14 & 20 & 44 & 71 & 61 & 78 \\
\bottomrule
\end{tabular}
\end{table*}

%% file: sec/evaluation_tasks.tex
\section{Testing AI's ability to Extract Model and Code From an Image}
The ability to form a representative model of the system beyond a given observation is a core aspect of true understanding. Applied to visual comprehension, this means going beyond surface-level pattern recognition to infer the underlying generative process behind an image. A direct way to evaluate this form of understanding is to test whether AI can analyze an image and identify, or even reconstruct, the system that produced it. We introduce three novel tasks to assess the AI’s  ability to link visual patterns to their underlying models. Each task was used to evaluate leading Vision Language Models (VLMs), and the results appear in \cref{tab:results}. The scripts used to run these tests are given in the SI.\newline
\textbf{Im2Code/Im2Desc (Image-to-Code Matching):} The most straightforward approach involves matching images to their generative code. The AI receives either the full code or verbal descriptions of generative processes as references, along with several test images, and must identify which set of images was created by the process  (\cref{fig:fig3}). For the results in \cref{tab:results} we supplied images from 10 different models with 3 reference images from each model. \newline
\textbf{Im2Im (Image-to-Image Model Matching):} This task provides the AI with a set of reference images, all generated by the same process, plus several test images. The AI must identify which test image was produced by the same process as the references (\cref{fig:fig3}). For the results in \cref{tab:results}, we supplied 3 reference images created by the same model, and 10 test images each created by a different model (\cref{fig:fig3}).\newline
\textbf{Im2Sim2Im (Image-to-Simulation-to-Image):} The most challenging test involves reconstructing the process beyond a real-world image (\cref{fig:fig4}.left). Given a natural photograph of a real-world phenomenon, the AI must identify the underlying physical process that created it, implement that process as executable code, and run the simulation to generate new images. Unlike the previous tasks, where the AI was required to select from predefined options, here the AI task is to reconstruct the model from scratch. The key challenge in this approach is evaluating the accuracy of AI-inferred models. Since the final output is an image generated by the model (\cref{fig:fig4}.left), we propose using visual similarity between the input natural image and the generated output image as a proxy for model accuracy. This approach builds on the observation that both AI and humans excel at the Im2Im matching task (\cref{fig:fig3}), reliably identifying images produced by the same process with high accuracy (\cref{tab:results}: Im2Im). We leverage this robust matching capability as an evaluation mechanism. We present the evaluator with the reference natural image alongside a test set containing the simulated image (generated by the inferred model) plus several distractor images from alternative models. The matcher identifies which test image best replicates the process that formed the reference image (\cref{fig:fig4}.left).  It's clear that the accuracy of this matching task should be correlated to the accuracy of the model that generates the image (Assuming an accurate evaluator). The rationale for the metric is: if a model accurately captures the underlying mechanism, it should produce images that match the reference better than images produced from alternative models (that were generated in response to different images). While the correlation between image similarity and model fidelity is imperfect, this limitation is inevitable. As far as we know, no universal metric exists for evaluating general-purpose models across arbitrary domains. The ability of a model to recreate observable outputs (in this case, visual patterns) provides a practical and intuitive measure of model quality (\cref{fig:fig5}). Hence, the ability of a model to generate a pattern that looks like the observed pattern counts as a strong indication of its validity [1-5]. We tested three type of evaluator/matchers for images similarity (\cref{fig:fig4}.left). Two based on GPT-5 with color or grayscale images (to isolate the impact of colors), and one based on a human preference. The results show relative consistency mostly within 10\% between the methods (\cref{tab:results}.Im2Sim2Im).

%% file: sec/result_and_conclusion.tex
\section{Results} 
The results of the tests are summarized in \cref{tab:results}. These results show that most large vision language models (VLMs) can successfully connect the images to the models and code that formed them, with performance substantially exceeding random chance (10\% \cref{tab:results}). All VLMs except Grok4 achieved high accuracy in matching different images generated by the same model (\cref{fig:fig3}), with GPT-5 and Gemini Pro 2.5 reaching accuracies up to 95\% (\cref{tab:results}). In the more challenging Im2Code task of matching images to both the model and the code that produced them (\cref{fig:fig3}), most VLMs achieved accuracies in the 40–60\% range. Notably, performance was comparable when image matching to code (Im2Code) and to model descriptions (Im2Desc) (\cref{tab:results}). This suggests that VLMs extract similar information from code and textual model descriptions. In the Im2Code task, we compared performance on matching images to the clean code (without comments) and annotated code (with comments). Both conditions yielded similar accuracy, indicating that comments provided no additional information to the VLMs. When tasked with identifying and replicating the generative process behind real-world images (Im2Sim2Im; \cref{fig:fig4,fig:fig5}), the VLMs produced impressive results (\cref{tab:results}). Showing VLM's ability to create models that captured key physical aspects of the real systems,  at least at a simplified toy model level (\cref{fig:fig5}).\newline
\textbf{What Does the AI See?} For all tests, the VLMs were asked to explain their answers. These explanations reveal that the VLMs mostly focus on inferring visual features (such as smooth transitions and scanline-like patterns) from the code and match them to features extracted from the images. While the visual features inferred from both code and images were mostly accurate, they tended to be general and often lacked distinctiveness, leading to matching errors.
\subsection{Im2Model2Code: From Image to Model to Code}  
When matching images to code, all VLMs appeared to prioritize inferring visual patterns from the code/model over extracting the underlying generative process from the image (see above). This suggests that the VLMs' approach to this task relies more heavily on their ability to infer visual features from code than on their ability to reverse-engineer the model from the image. To address this limitation, we redesigned the test with a two-stage process. First, one VLM was asked to examine each image and infer the generative process/model behind it. These textual descriptions were then provided (without the images) to a second  LLM, which was asked to match the described systems to the corresponding code. This approach forces the first model to extract the underlying model from the image, rather than simply inferring visual patterns from code, and the second LLM to use only model description and not image patterns for matching. As shown in \cref{tab:results} (Im2Model2Code vs Im2Code), this two-stage approach yielded lower accuracy but still performed significantly above random chance. This indicates that most VLMs possess a limited but clear ability to infer models from images. We also inspected a few inferred models to qualitatively validate this and found that, for simple cases, these models matched the physical system in the image.\newline
\subsection{Im2Sim2Im: Results Extracting Model From Image}
The Im2Sim2Im test involves giving the AI an image of a real-world pattern and asking it to identify the process that forms this pattern,  implement it in code, and run the code to generate a simulated image (\cref{fig:fig4}.left). This task required creating models and code, with only an image as a reference, making it considerably more challenging than previous tests.  Most VLMs easily identified the visual patterns and their physical sources. This is not surprising since the test images depicted well-known phenomena (foam, dust, clouds, plants, cities; \cref{fig:fig5}).  Accurately simulating many of these patterns is complex and often computationally infeasible. Consequently, in the majority of cases, the VLMs generated simplified toy models rather than accurate simulations. The relatively good scores VLMs got in this test (\cref{tab:results}:Im2Sim2Im) imply that  the generated models capture the main essence of the system (\cref{fig:fig5}).  A notable result was the ability of leading VLMs  to deal with complex systems by representing each of their components in different levels of abstraction.  For example, when presented with an image of sunlight reflecting on water waves (\cref{fig:fig4}), GPT-5 represented the waves as a simple sinusoidal function, then used the surface normals with Snell's law to calculate light reflections and refractions from both the water surface and pool floor (\cref{fig:fig4}). This demonstrates an impressive capacity to integrate crude approximations (sinusoids for waves) with real physics (Snell's law) to model complex systems using only a single reference image. This  ability to represent different elements of the system in different levels of abstraction is a sign of deep understanding. While the generated images were crude (\cref{fig:fig5}), they captured essential system characteristics and demonstrated that the VLMs understood both the patterns and their underlying physics. For instance, when presented with a printed book page, GPT simulated a Markovian chain of letters modeling letter distribution (\cref{fig:fig5}). However, when given an image of a handwritten page (\cref{fig:fig5}), the AI simulated the brush stroke, replicating the curve structure but losing the letter meaning, again showing the VLM's ability to simulate a similar system at different levels of abstraction. Smaller VLMs, such as GPT-5-mini and Gemini-2.5-Flash, often produced overly simplistic but more tunable models.    Interestingly, these simplified models were frequently better tuned to the input image and more successfully replicated pattern appearance compared to more physically accurate simulations of the same patterns made by larger VLMs. Whether less accurate simulations that are more tunable and hence can be better fitted to the observed pattern can be considered  superior to more physically accurate but less tunable models is an open question in science\cite{rushkin2015optimizing}.
The evaluation of image similarity (\cref{fig:fig4}.left) was done using  either AI (GPT-5) or human preference, with the results of both approaches falling mostly within 10\% range (\cref{tab:results}.Im2sim2Im). Removing the color and matching only grayscale images lead to around 10\% drop in accuracy (\cref{tab:results} gray/color), implying that color is significant but not critical to the image matching.  The AI image matcher (\cref{fig:fig4}.left) used both low-level features (strips/textures) and semantic content (letters, scene, trees) when identifying similar images, again showing an ability to understand the image across multiple levels of abstraction (based on  the VLM's own explanations).

\section{Conclusion}
This work leverages recent advances in Agentic AI to tackle two fundamental challenges in visual understanding: the formation of visual patterns and the modeling of their underlying mechanisms. Our first contribution is a large-scale dataset of visual patterns and textures spanning multiple domains of science, technology, and art. Critically, the dataset includes not only images but also the models and code that generate them, enabling deep exploration of the relationship between visual patterns and the processes that produce them. In addition we explore the ability of AI agents to invent novel generative models for textures and patterns generation, inspire either by natural images or by the LLM original ideas.  In addition, for being significant resource for any field that relies on textures and visual patterns \cite{ambientcg,polyhaven,drehwald2023one,cimpoi2014describing,sharma2023materialistic,raistrick2023infinite}, the dataset opens the door to general exploration of the deep connection between visual patterns and the mechanisms that form them. 
The availability of paired images and their generative code enables a fundamentally new approach to evaluating visual understanding. Rather than assessing only pattern recognition, we can now probe whether AI systems comprehend the mechanisms that produce visual phenomena. 
Understanding the processes underlying visual patterns, rather than merely recognizing them, is a core aspect of genuine comprehension, yet remains largely unexplored in computer vision. The dataset and the three novel evaluation methods presented here allow us to assess vision-language models' capacity to infer the systems behind observed visual patterns in both qualitative and quantitative ways. The results clearly demonstrate that state-of-the-art VLMs can infer physical mechanisms and models from images alone. This capability transcends traditional image classification and pattern recognition, marking a significant step toward general visual intelligence. Developing methods to explore, harness, and enhance this mechanistic understanding is a central challenge for future research in computer vision and AI.